\documentclass[10pt,twocolumn,letterpaper]{article}

\usepackage{cvpr}
\usepackage{times}
\usepackage{epsfig}
\usepackage{subfig}
\usepackage{graphicx}
\usepackage{amsmath}
\usepackage{amssymb}

\usepackage[pagebackref=true,breaklinks=true,letterpaper=true,colorlinks,bookmarks=false]{hyperref}

\cvprfinalcopy 

\def\cvprPaperID{xxx} 

\ifcvprfinal\fi
\begin{document}
\title{Reflectance Hashing for Material Recognition}

\author{Hang Zhang\\
Rutgers University\\
94 Brett Road, \\
Piscataway,\\
NJ, 08854.\\
{\tt\small zhang.hang@rutgers.edu}
\and
Kristin Dana\\
Rutgers University\\
94 Brett Road, \\
Piscataway,\\
NJ, 08854.\\
{\tt\small kdana@ece.rutgers.edu}
\and
Ko  Nishino\\
Drexel University\\
3141 Chestnut Street,\\
Philadelphia,\\
PA 19104.\\
{\tt\small kon@drexel.edu}
}

\maketitle



\begin{abstract}
  
  We introduce a novel method for using reflectance to identify materials. Reflectance offers a unique signature of the material but  is challenging to measure and use for recognizing materials due to its high-dimensionality.  In this work,  one-shot reflectance is captured using a unique optical camera measuring   {\it reflectance disks} where the pixel coordinates correspond to surface viewing angles.  The reflectance has class-specific stucture and angular gradients computed in this  reflectance space  reveal the material class.  
  These reflectance disks encode discriminative information for efficient and accurate material recognition. We introduce a framework called reflectance hashing that models the reflectance disks with dictionary learning and binary hashing.  We demonstrate  the effectiveness of reflectance hashing for material recognition with a number of real-world materials.
  
\end{abstract}

\section{Introduction}
\begin{figure}[t]
\begin{center}
\includegraphics[width=2.5in]{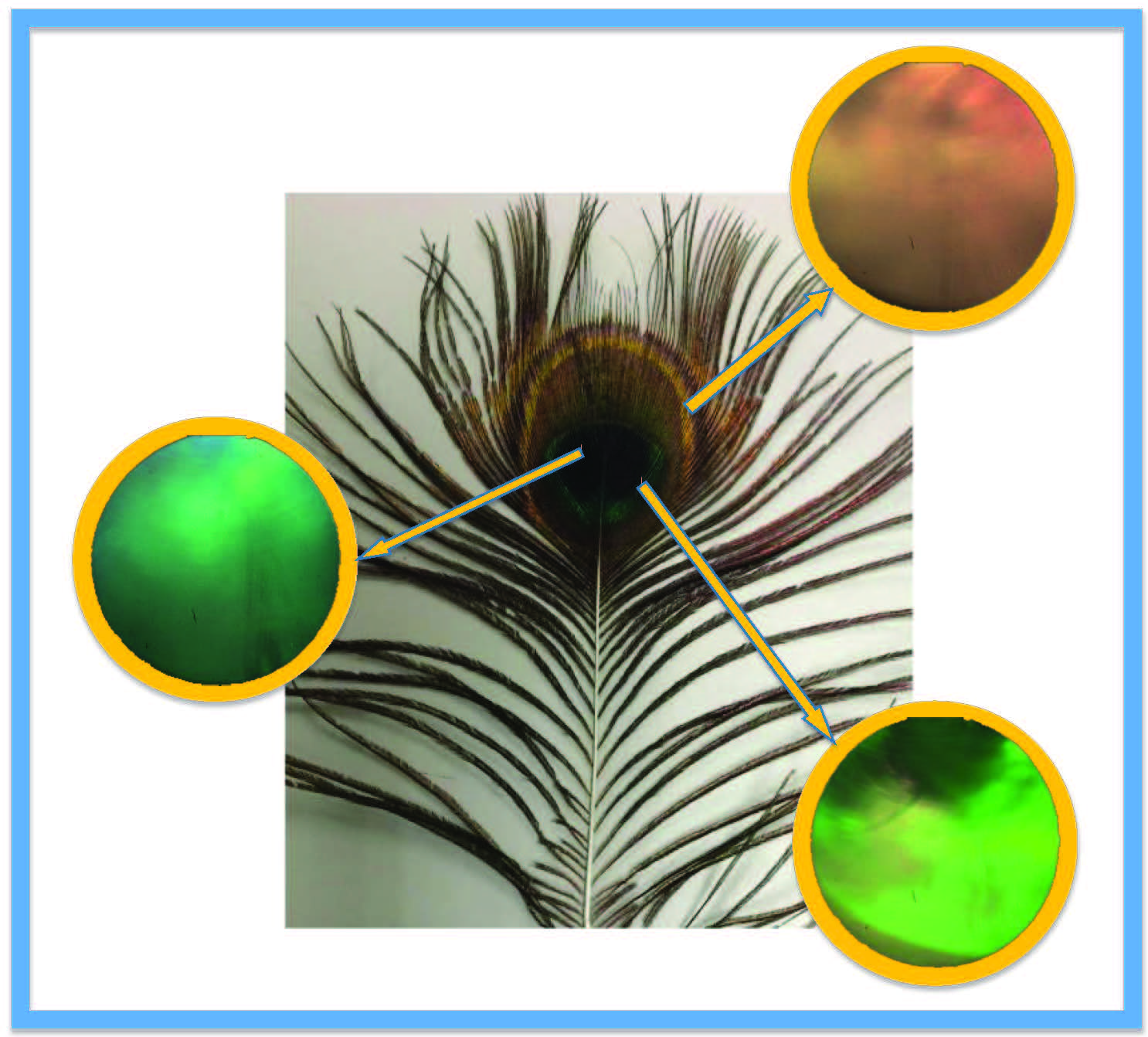}
\end{center}
   \caption{ Reflectance disks provide a snapshot of the reflectance function for spot samples on the surface.  For this example of a peacock feather,  iridescence causes a large variation
   of intensity with viewing direction as revealed in the three reflectance disks. }
\label{fig:peacockfeather}
\end{figure}

 \vspace{-0.07in}
Color and geometry are not a full measure of the richness of visual appearance. Material composition of a physical surface point determines the characteristics of light interaction and the reflection to an observer.  In the everyday real world there are a vast number of materials that are useful to discern including concrete, metal, plastic, velvet, satin, asphalt, carpet, tile, skin, hair, wood, and marble. A computational method for identifying these materials has important implications in developing new algorithms and new technologies for a broad set of application domains. For example, a mobile robot or autonomous automobile can use material recognition to determine whether the terrain is asphalt, grass, gravel, ice or snow in order to optimize mechanical control.  An indoor mobile robot can distinguish among wood, tile, or carpet for cleaning tasks. The material composition of objects can be tagged for an e-commerce inventory or for characterizing multi-composite 3D printed objects. The potential applications are limitless in areas such as robotics, digital architecture, human-computer interaction, intelligent vehicles, and advanced manufacturing. Furthermore, just as computer vision algorithms now use depth sensors directly from RGB-D cameras, material sensors can have foundational importance in nearly all vision algorithms including segmentation, feature matching, scene recognition, image-based rendering, context-based search, object recognition, and motion estimation.
\begin{figure}[t]
\begin{center}
\subfloat[Leather Surface]
{
\includegraphics[width=0.285\linewidth]{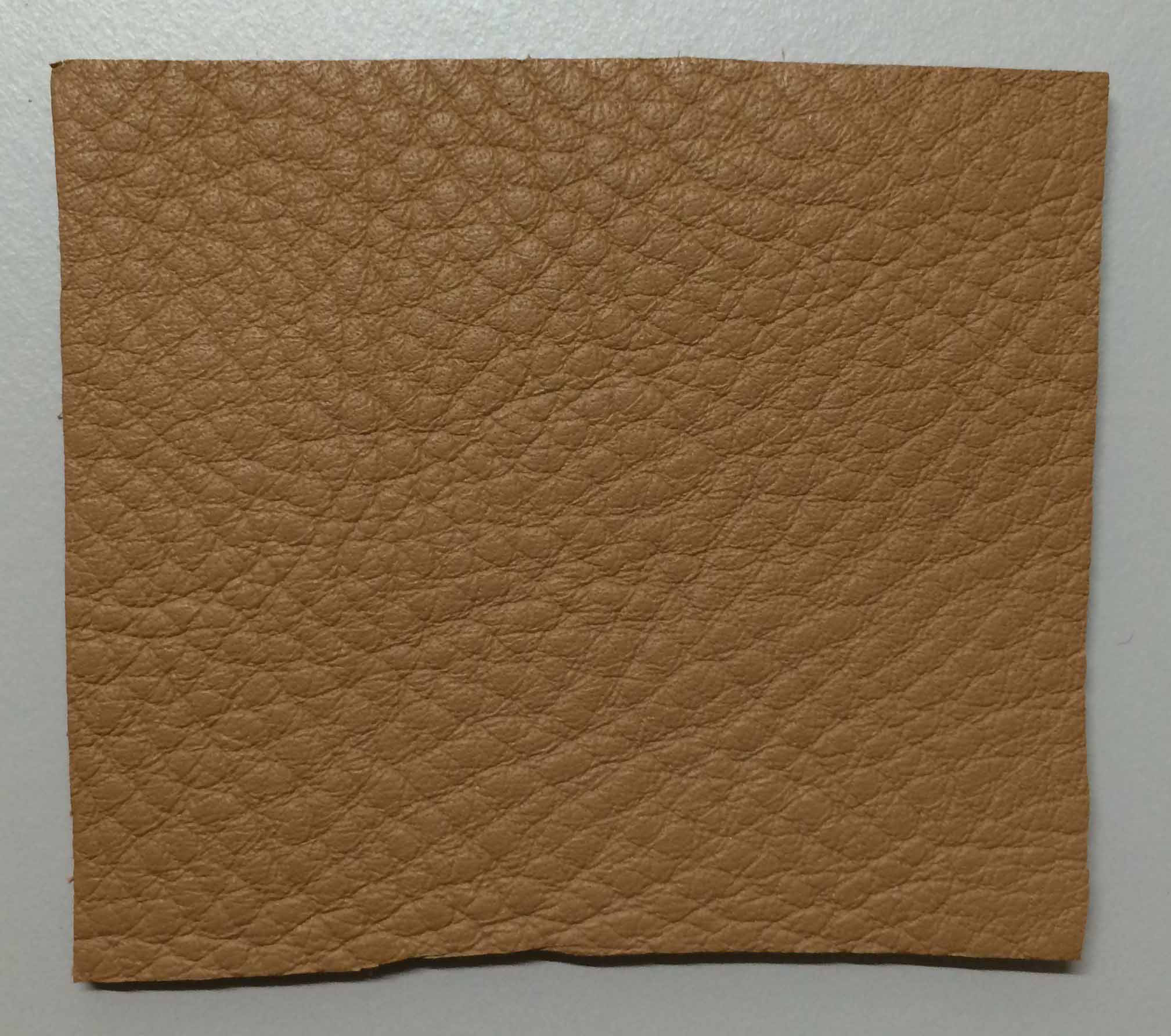}
}
\subfloat[Reflectance Disk]
{
\includegraphics[width=0.33\linewidth]{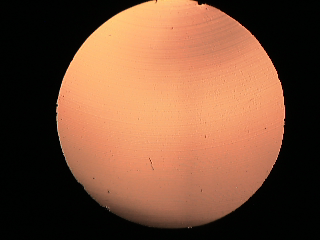}
}
\subfloat[Texton Map]
{
\includegraphics[width=0.33\linewidth]{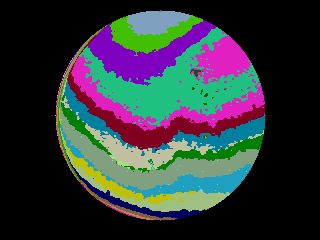}
}\\
\subfloat[Auto Paint Surface]
{
\includegraphics[width=0.285\linewidth,height=0.245\linewidth]{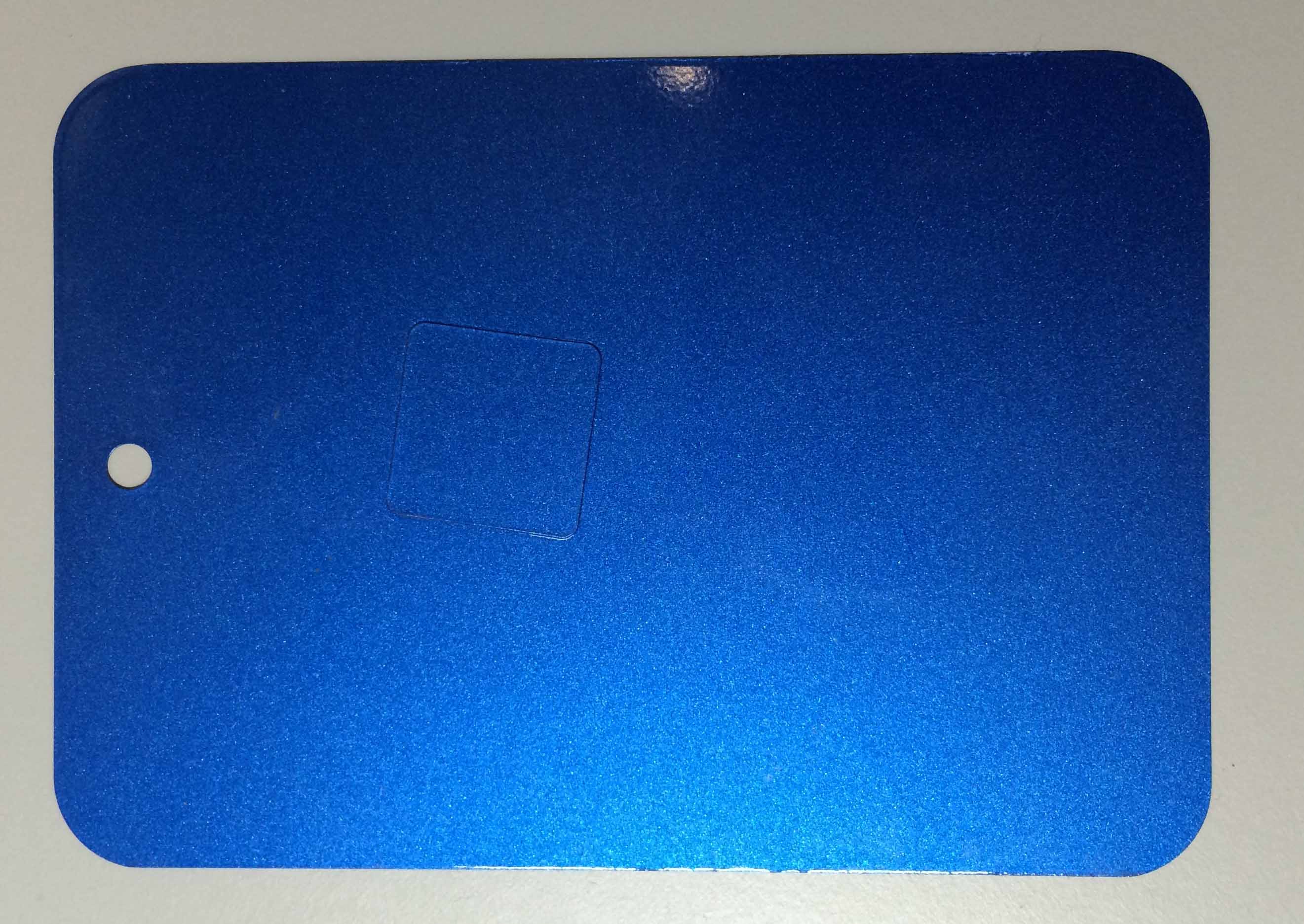}
}
\subfloat[Reflectance Disk]
{
\includegraphics[width=0.33\linewidth]{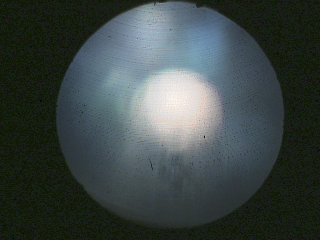}
}
\subfloat[Texton Map]
{
\includegraphics[width=0.33\linewidth]{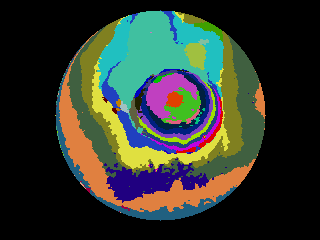}
}
\end{center}
\caption{Reflectance disks  provide a quick snapshot of the intrinsic reflectance of a surface point. Gradients of the reflectance space are captured with textons and 
provide a signature for material recognition.  }
\label{fig:reflectancedisk}
\end{figure}
Our approach uses reflectance for material recognition because of the
advantages of having an intrinsic optical signature for the surface. However,  we bypass the use of a gonioreflectometer by using a novel one-shot reflectance camera based on a parabolic mirror design.  The output of this camera is a {\it reflectance disk}, a dense sampling of the surface reflectance of the material projected into a single image as shown for two example surfaces in Figure~\ref{fig:reflectancedisk}. Each reflectance disk measures surface properties for a single point and can capture complex appearance such as the iridescence of a peacock feather as illustrated in Figure~\ref{fig:peacockfeather}. 
We then use this convenient reflectance measurement as the discriminative characteristic for recognizing the material class.  We address the issue of high dimensionality using a 
novel application of binary hash codes to encode reflectance information in an efficient yet discriminitve representation. 
The key idea is to obtain sufficient sampling of
the reflectance with enough discriminative power and reduce its
representation size so that it can be effectively used for probing the
material.

We present a database of reflectance images comprised of twenty different diverse material classes including  wood, velvet, ceramic and automotive paint with 10 spot measurements per surface and  with three different surface instances per class. Measurements include
three on-axis illumination angles ($-10^\circ,0^\circ, 10^\circ$) and ten random spot measurements over the surface.  Each spot measurement is a reflectance disk composed of a dense sampling of viewing angles totaling thousands of reflectance angles per disk.
The  database of 3600 images or reflectance disks is made publicly available.   
For recognition, we combine binary hash coding  and texton boosting for a new framework called {\it reflectance hashing}  for efficient and accurate recognition  of materials.   
 We compare  reflectance hashing with  texton boosting for the task of recognizing  materials from reflectance disks. 


\section{Related work}
 \vspace{-0.07in}
Prior methods for material recognition use two distinct approaches.  One approach  assesses material identity using reflectance as an intrinsic property of the surface  \cite{Lombardi12a,Cula04a,Leung01,Kumar10,Gu12}.  
Another main approach identifies material labels using the appearance of the surface within the real world scene \cite{Sharan13,Liu10, Li12}. 
Using reflectance instead of scene appearance has the advantage that reflectance is
a direct measurement of the material characteristics, instead of its
phenomenological appearance  \cite{Adelson01}. Reflectance is mostly unique to the material, whereas the
appearance is the convoluted end result of the interaction of all the
intrinsic and extrinsic factors and thus more difficult to decipher.
 
 \begin{figure}[t]
\begin{center}
\includegraphics[width=3in]{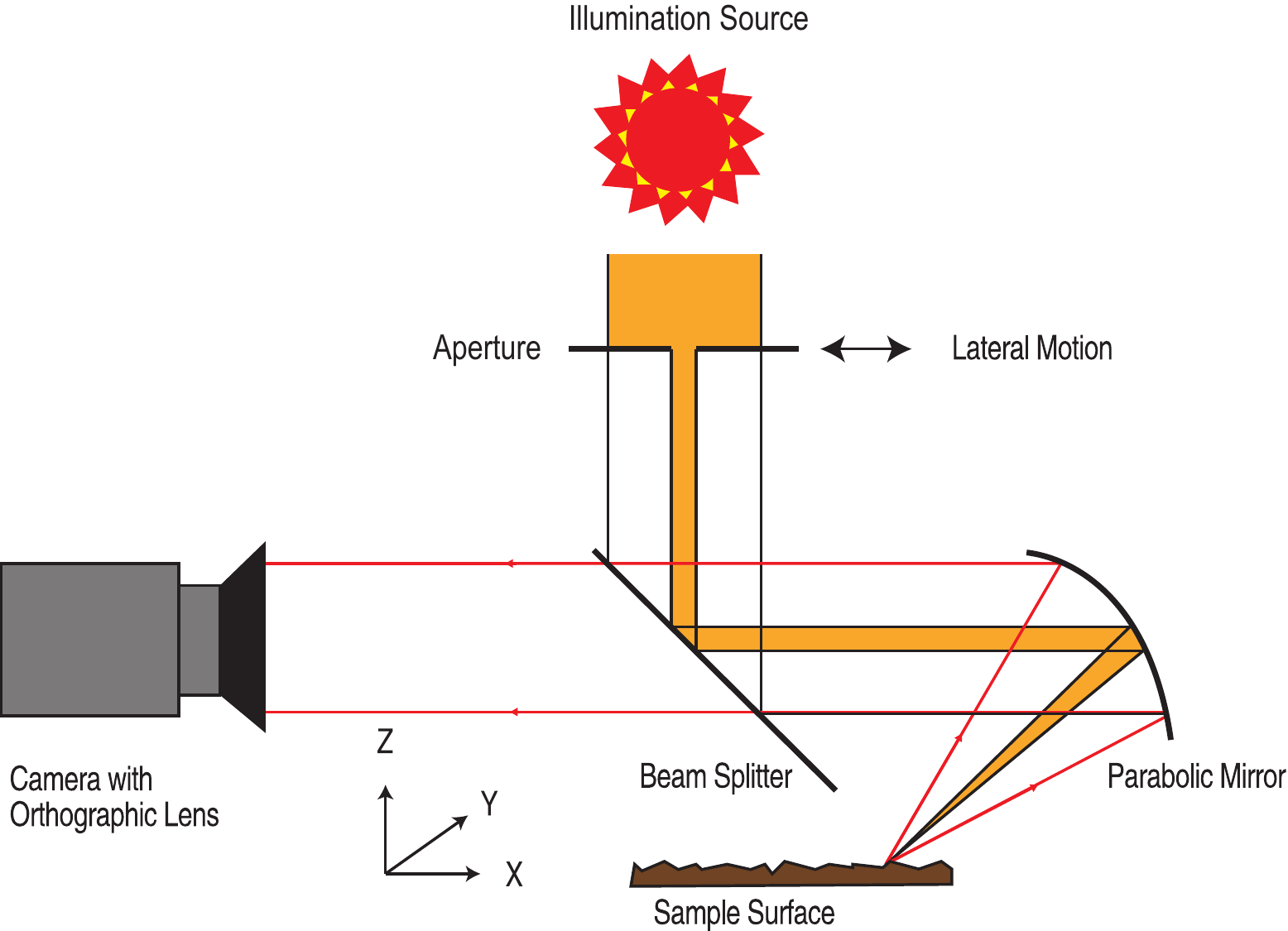}
\end{center}
\caption{Schematic of the mirror-based camera. Reflectance disks are obtained by viewing a single point under multiple viewing directions using a concave parabolic mirror viewed by a telecentric lens.}
\label{fig:cameraschematic}
\end{figure}

\begin{figure*}[t]
\begin{center}
\includegraphics[width=4.9in]{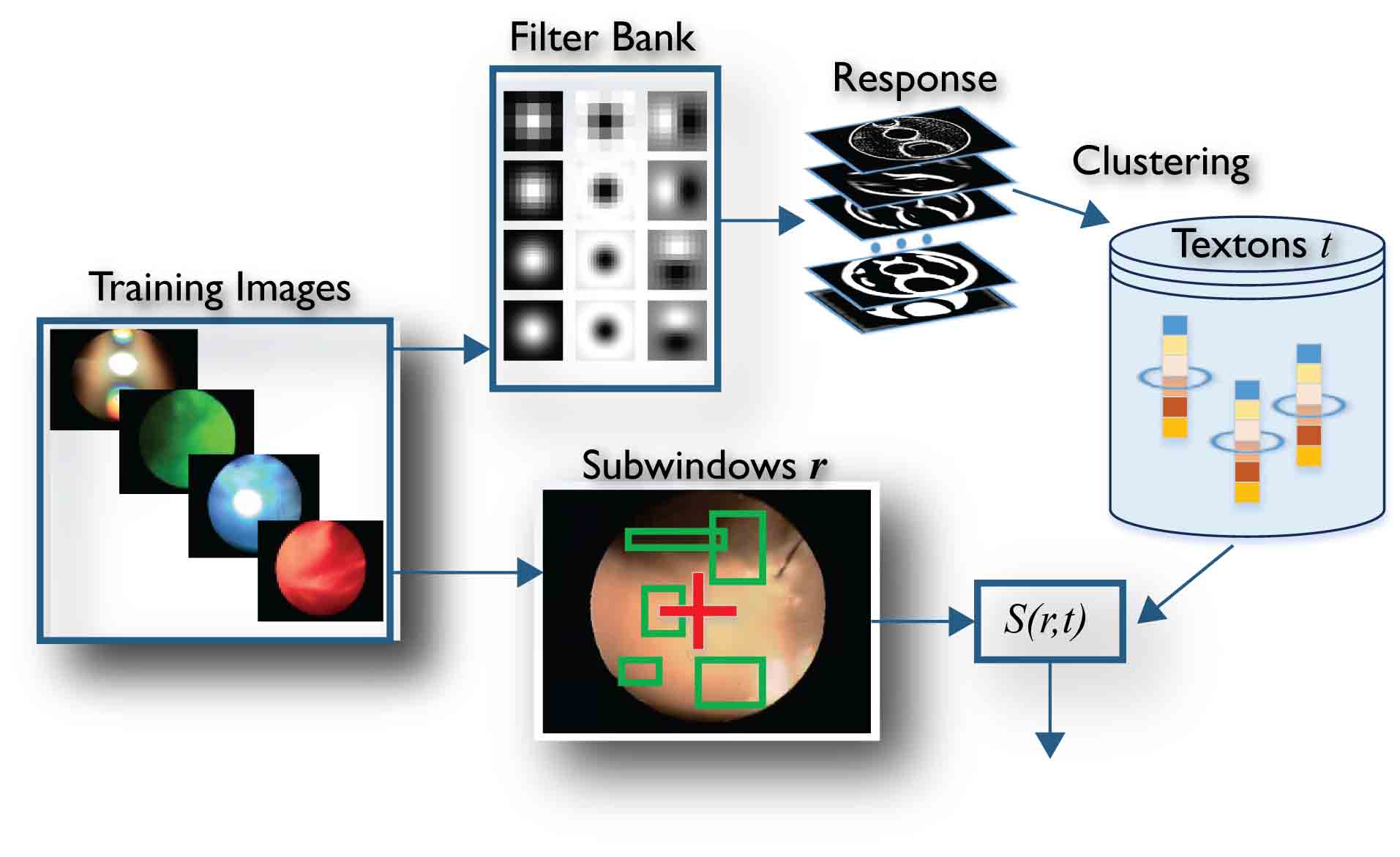}
\end{center}
\caption{Gradients on the Reflectance Disk Extracted with Textons: Images are filtered by a  filter bank comprised of Gaussian, Laplacian and oriented gradient filters. The  $24 \times 1$ responses  vectors are clustered to form visual words or textons.  A random subsampling of rectangular subregions of the reflectance space defines regions of interest $r$. The feature of interest is $S(r,t)$, the count of pixels identified as  texton $t$ in region $r$.  
}
\label{fig:textonmethod}
\end{figure*}

A challenge in using reflectance for
material recognition is that measurements are typically difficult and cumbersome. For example, most methods require
knowledge of the scene like geometry \cite{Lombardi12a, Lombardi12b} or illumination \cite{Oxholm14,Oxholm12}.  Other methods require
  lab-based measurements  of the BRDF (bidirectional reflectance distribution function) or BRDF slices such as light domes for illumination patterns \cite{Liu14}.
Acquiring full BRDF requires gonioreflectometers that are comprised of multiple cameras and light sources
covering the hemisphere of possible directions using geodesic domes or robotics as in \cite{Foo96,Ward92,Levoy96, Dana99,Matusik03,Matusik_measure03, Wang_CVPR09}.  
Surface appearance in terms of relief texture is captured by the GelSight camera \cite{Johnson09} using three-color photometric stereo by imprinting the surface geometry on an elastomer surface with Lambertian skin. With this device very fine geometric texture can be recovered but surface appearance is lost. 
 An additional challenge of  using reflectance for  measurements is the high dimensionality, reflectance as a function of illumination and viewing direction especially if densely sampled can lead to thousands or more samples per surface point.

Specialized cameras have been developed in prior research to obtain reflectance measurements of surfaces efficiently.  The mirror-based camera illustrated in Figure~\ref{fig:texcam} is one  such a device \cite{Dana99,Dana04}.  This camera views an image of a single surface point by using a concave parabolic mirror with the focus coincident with the target point. In this manner, the camera records a multiview image of a surface point where each pixel records a different angle.  Multiple views from mirror-based optics is also achieved with a kaleidoscope-like configuration \cite{Han03}. However,  the viewing angles captured in this device are discrete and sparse. Because our approach relies uses gradients in angular space a dense sampling of reflectance is needed and   we use the parabolic mirror-based camera as a reflectance sensor.

 In prior work,  recognition of standard scene images is typically accomplished using image features that capture the spatial variation of image intensity. For example, image filtering with a set of filter banks  followed by grouping the outputs into characteristic textons has been used  for  image texture \cite{Leung01,Cula01a,Varma09}, object recognition \cite{Shotton06,Shotton09} and image segmentation \cite{Ladicky14}. 
 Similarly, we encode the discriminative optical characteristics of materials captured in the reflectance disks with a texton-based representation.
Texton methods for the general problem of scene and image recognition have been improved significantly by incorporating 
 boosting  as in 
 {\it Texton-boost} \cite{Shotton09}  and {\it Joint-boost} \cite{Torralba07}. We incorporate the utility of boosting to identify  a discriminitive  and compact representation. 
 
The goal of efficient recognition of images for large scale search has led to numerous methods for binary hash codes \cite{Kulius09,Bergamo11,Wang10,Liu11,Liu12}.
A common approach in recent work \cite{Weiss08,Torralba08,Kulius09,Raginsky10} is  learning  binary codes where the Hamming distance approximates the kernel distance between appearance descriptors. This method is used when the  kernel distance is known to give good performance but is expensive to compute and store.
Inspired by the work in image search, 
we  develop {\it reflectance hashing} with binary codes that allow matching of measured reflectance disks with those in the labeled training set.  The binary codes are learned using the match metric obtained from the texton-based descriptor we present in Section~\ref{sec:methods}.  We use 
a suite of state-of-the-art  hashing techniques applied to our measured reflectance disk dataset  to 
evaluate recognition performance  in terms of speed and recognition accuracy.

%
%
\section{Methods}
\label{sec:methods}
\subsection{Reflectance Disk Measurements}
%
We use the {\it texture camera} approach introduced in \cite{Dana01,Dana04} for  measuring reflectance of surface points, 
The measurement camera mainly consists of a parabolic mirror, a CCD camera, a movable aperture and a beam splitter. The imaging components and their arrangement are illustrated in Figure~\ref{fig:cameraschematic}. 
 A parabolic mirror section is fixed so that its focus is at the surface point to be measured. The illumination source is a collimated beam of light parallel to the global plane of the surface and passing through a movable aperture. The angle of the incident ray reaching the surface is determined by the intersection point with the mirror.  Therefore, the illumination 
 direction is determined by  the aperture position.  Light reflected  from the surface point is reflected by the mirror to a set of  parallel rays directed through a beam splitter to the camera.  Each pixel of this camera image corresponds to a viewing direction of the surface point.  Therefore, the recorded image is an observation
 of  a single point on the surface but from a dense sampling of viewing directions.  Multiple illumination directions
can be measured by planar motions of  the illumination aperture. 
A key advantage of this approach is the potential  for a handheld reflectance sensor where the mirror and light source are attached to a handheld camera.

We refer to the camera's output images as {\it reflectance disks} since they depict circular regions  from  on-axis projections of a paraboloid.  Examples of these  reflectance disks are shown in Figure~\ref{fig:reflectancedisk} for two surfaces.  These examples show the variation of the surface reflectance with viewing angle. 
Reflectance disks have an intensity variation that corresponds to angular change with the viewing direction instead of spatial change on the surface.  However, the reflectance disks can be interpreted as images and  filtered to find characteristic gradients.   This approach of using {\it angular gradients in reflectance space} 
 is a   novel contribution of our work.
Our goal is to represent the measurements in an efficient and meaningful way that supports material recognition.



\subsection{Textons on the Reflectance Disk}
\begin{figure}[t]
\begin{center}
\includegraphics[width=2.7in]{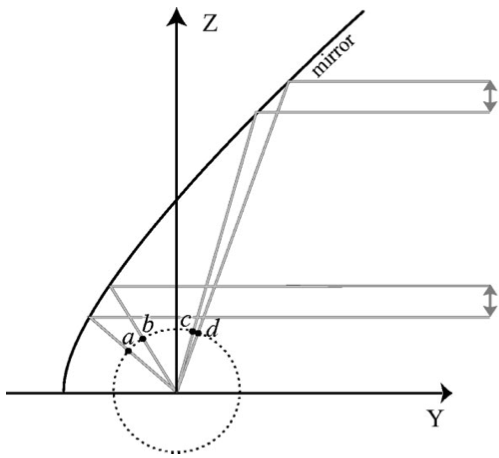}
\end{center}
   \caption{ Gradients on the reflectance disk are computed using image filtering. However, the angular resolution varies over the surface as shown above. The angle at $ab$ is different from $cd$. Therefore differences (or filtering) computed with a uniform spacing on the reflectance disk have a non-uniform mapping to angles.  }
\label{fig:anglevariation}
\end{figure}
We find characteristic patterns of intensity in a reflectance disk by 
 computing {\it spatial change} in the reflectance disk via image filtering. 
 Since spatial coordinates of the reflectance disk map to angular coordinates of reflectance, filtering is a convenient way to compute  angular gradients of the reflectance.  
 A common method of quantifying intensity patterns is using {\it textons}. Textons are
 computed by first filtering the image with a set of gradient filters at different orientations
 and scales. 
 We employ the same filter banks as in \cite{Leung01},  comprised of  Gaussian low-pass filters at four scales, Laplacian filters at four scales and eight gradient filters at different orientations and scales for a total of 24 filters.  The resulting $24 \times 1$ responses at each pixel are clustered using   $K$-means clustering and a  dictionary of visual words called textons is created.  The underlying assumption is that the local intensity variation can be captured with the finite set of characteristic filter responses that are the centers of each of the $K$ clusters. 

We use textons as a reflectance feature to provide a dense, per-pixel description of reflectance variation.  
Methods to detect key-points are useful for scene images, but reflectance disks do not typically have specific key-points of interest.  

For texton computation, we use  gradient filters that approximate the derivative operator with a discrete spatial difference filter. 
This discrete approximation of the gradient  computes change over a non-infinitesimal distance, typically the difference of neighboring pixels.  The distance between these neighboring pixels  is constant over the reflectance disk, i.e.\ the same gradient filter is applied over the entire disk image. However,  because of the mapping from spatial coordinates to angular coordinates by the parabolic mirror, the angular distance is not constant. This situation is illustrated in Figure~\ref{fig:anglevariation} where the cone angle at the surface is shown for two different locations on the parabolic mirror. A constant distance $d$ on the projected disk leads to two different cone angles. 
Consequently,  the resolution for the angular gradient  varies as a function of disk coordinates. 

The equation of the  parabolic mirror surface is given by, 
\begin{equation}
y+F = \frac{ z^2 + x^2}{4F}, 
\end{equation}
where the $y$-axis is the optical axis of the mirror aligned with the camera optical axis, 
the $x-z$ plane is aligned with the disk-shaped projection of the mirror surface on camera's image plane, and  $F$ is the mirror focal length (12.7 mm).  Define the cone angle $\alpha$ as the set of angles subtended by arc $ab$ or $cd$ in Figure~\ref{fig:anglevariation}, where $d$ is taken as 2 pixel units. 
From this equation, the variation of the cone angle $\alpha$  as a function of the vertical spatial coordinate $z$ on the reflectance disk is derived in \cite{Dana04} as 
\begin{equation}
\alpha(z) = \arctan \frac {z-1}{\frac{(z-1)^2}{4F} - F} - \arctan \frac {z+1}{\frac{(z+1)^2}{4F} - F}.
\end{equation}
Since the geometry of the mirror is known, the spatial filters can be adaptive to ensure
a uniform  cone angle over the disk.  However,  the spatially invariant gradient filter provides a discriminitive representation and uniformity of
angular resolution is not required.


\paragraph{\bf Reflectance-Layout Filter}
Textons are a local measure of image structure, and texton histograms are used to represent a region.  The histogram representation has the advantage of invariance to shifts of structures. 
 Reflectance disks show the structure of the reflectance field, but often the distribution within that structure is the characteristic property. 
Histogram representations have the drawback that crucial  spatial information is lost.  We follow the texton boost framework \cite{Shotton06,Shotton09,Torralba07} to encode
the characteristic spatial structure of the reflectance disk by defining  a set of randomly shaped rectangular regions over the disk.  
For a given texton $t$ and region $r$, $S(r,t)$ is the count of pixels labelled with texton $t$ that are located within region $r$ as illustrated in Figure~\ref{fig:textonmethod}.  We use this response as a key feature for describing the reflectance disks. This response is referred to as the texture-layout filter response when applied to ordinary images \cite{Shotton09}. For material recognition using reflectance disks, we use the term {\it reflectance-layout filter} to indicate that the representation captures the angular layout of reflectance gradients. 
For computing reflectance-layout filter responses, the number of particular textons in a region, the texton count 
 is calculated using a soft-weighting for the 8 nearest cluster centers.  
 Soft-weighted textons as in \cite{Gemert08,Shotton09,Ladicky14} allows each pixel  to be associated with 8 nearest clusters instead of a single texton label. This method alleviates the problem of two similar responses assigned to different texton labels. 

Each reflectance layout filter characterizes a specific region in reflectance space where the 
change of reflectance as a function of viewing angle is a descriptive feature for the material.
There are many possible combinations of $r,t$ as illustrated in Figure~\ref{fig:textonmethod} and we want to choose the ones that are most discriminitave. The boosting approach in \cite{Shotton06,Torralba07} is used to identify the most discriminitive $r,t$ combinations.  
After boosting, the typical numbers of reflectance layout filters is approximately 700. 
\subsection{Boosting}

The  algorithm for {\it Joint Boosting}  \cite{Torralba07} provides  feature selection where the feature is the reflectance-layout filter $S(r,t)$ specified by a particular $(r,t)$ pair that indicates the number of times texton $t$ occurs in region $r$.
The approach builds a strong learner iteratively by adding a strongest weak learner which is the selected feature in each iteration, allowing the weak learner to help classify a subset of classes.   At each step $m$, the method maximally reduces the weighted squared error by choosing a strongest weak learner $h_m(I_i,c)$, updating strong learner as: $H(I_i,c):=H(I_i,c)+h_m(I_i,c)$, where $I_i$ is the input image and $c$ is the class. The weighted square error is given by
\begin{equation}
\label{eq:wse}
J_{wse}=\sum_{c=1}^C \sum_{i=1}^N w_i^c(z_i^c-h_m(I_i,c))^2
\end{equation}
where $C$ is number of classes, and $N$ is the number of training samples, $w_i^c=e^{-z_i^c H(I_i,c)}$ defines the weights for class $c$ and sample $i$ that emphasize the incorrectly classified samples for that iteration, $z_i^c$ is the membership label $(\pm1)$ for class $c$. The optimal solution that gives the selected feature at step $m$ has the form
\begin{equation}
h_m(c)=\left\{\begin{aligned}
&a\delta(S(r,t)>\theta)+b &\text{if } c\in T_m\\
&k^c &\text{otherwise}
\end{aligned}
\right.
\end{equation}
with the parameters $(a,b,k^c,\theta)$ given by the minimization of  Equation~\ref{eq:wse}, $\delta(.)$ is  indicator function (0 or 1), $S(r,t)$ is the count or texton $t$ in region $r$, $T_m$ is the class subset that share this feature. For the classes sharing the feature $(c_i\in T_m)$, the weak learner gives $h(I_i,c)\in\{a+b,b\}$ determined by the comparison of reflectance-layout filter response $S(r,t)$ with the threshold $\theta$. For the classes not sharing the feature, there is a constant $k^c$ different for each class $c$ that avoids the asymmetry caused by the number of positive and negative samples for each class.

This method chooses a dominant feature $h_m(I_i,c)$ at each iteration to minimize the error function. For each selected feature,  finding the class subset $T$ that maximally reduces the error on weighted training set is expensive. Instead of searching all possible $2^C-1$ combinations, a greedy search heuristic \cite{Torralba07} is used to reduce the complexity to $O(C^2)$. We first find the best feature for each single class, and pick the class that has the best error reduction. Then we add the class that has best joint error reduction until we go through all the classes. Finally, we select the set that provides the largest overall error reduction.
It is also time consuming  to go through all the features at each iteration,  so  only a fraction $\tau \ll 1$ reflectance-layout filters are examined, randomly chosen in each iteration. 


\subsection{ From Textons to Binary Material Codes}
\label{sec:hashing}
Boosting allows the selection of features $h_m(I_i,c)$ that specify $r,t$  (region, texton) pairs of interest.  The selected features can simply be viewed as the input feature vector for a basic nearest neighbor classification. Nearest neighbors in high dimensions is problematic, but the computational tools of binary hash codes enables efficient and accurate representation.
We use the reflectance layout filters directly in a concatenated feature vector with $700\times c$ component where each component is a region-texton $r,t$ pair that is known to be useful in material recognition.  There is likely to be redundancy in the $r,t$ combinations so the number of components in the feature vector may be reduced accordingly. 
This high dimensional feature vector is used directly for fast computation by employing binary coding.  This efficiency is highly desirable for the embedded hardware implementation.

We use a suite of state-of-the-art approaches for forming binary code words including  
circulant binary embedding (CBE) \cite{Yu14}, bilinear embedding \cite{Gong13a}, iterative quantization (ITQ) \cite{Gong13b},  angular quantization-based
 binary codes(AQBC) \cite{Gong12}, locality sensitive hash(LSH) \cite{Charikar02}, and locality-sensitive binary codes from shift-invariant kernels (SKLSH) \cite{Raginsky09}.  

\begin{figure*}
\begin{center}
\subfloat
{ 
\includegraphics[width=0.12\linewidth]{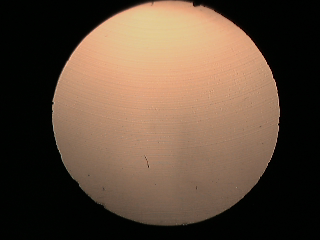}
}
\subfloat
{ 
\includegraphics[width=0.12\linewidth]{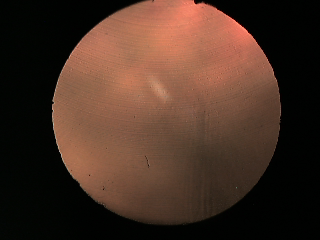}
}
\subfloat
{ 
\includegraphics[width=0.12\linewidth]{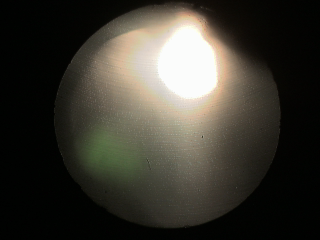}
}
\subfloat
{ 
\includegraphics[width=0.12\linewidth]{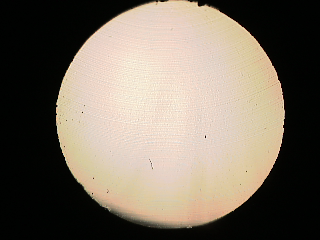}
}
\subfloat
{ 
\includegraphics[width=0.12\linewidth]{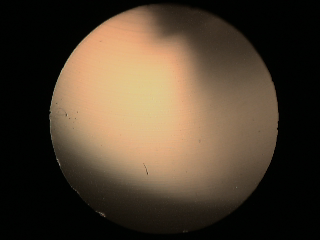}
}
\subfloat
{ 
\includegraphics[width=0.12\linewidth]{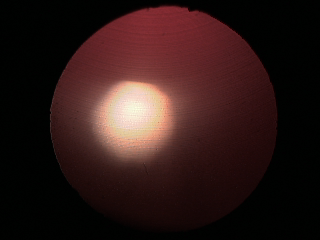}
}
\subfloat
{ 
\includegraphics[width=0.12\linewidth]{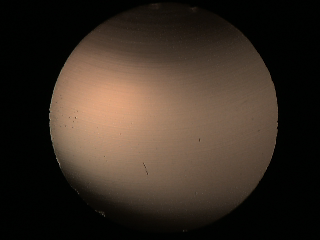}
}
\subfloat
{ 
\includegraphics[width=0.12\linewidth]{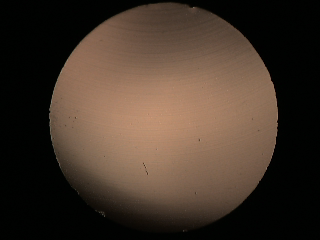}
}
\par
\subfloat
{ 
\includegraphics[width=0.12\linewidth]{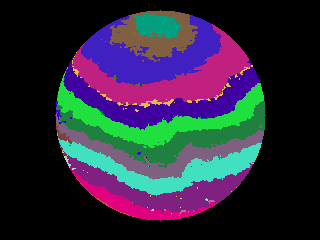}
}
\subfloat
{ 
\includegraphics[width=0.12\linewidth]{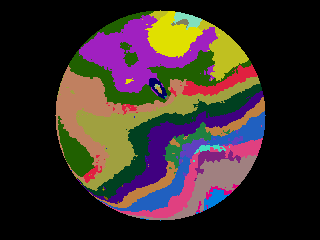}
}
\subfloat
{ 
\includegraphics[width=0.12\linewidth]{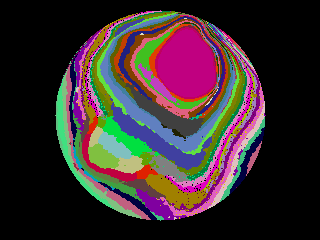}
}
\subfloat
{ 
\includegraphics[width=0.12\linewidth]{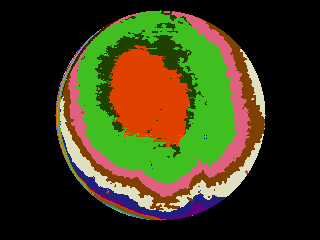}
}
\subfloat
{ 
\includegraphics[width=0.12\linewidth]{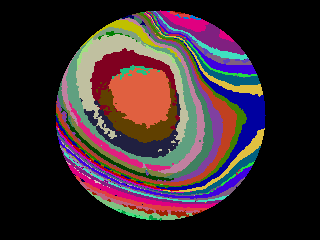}
}
\subfloat
{ 
\includegraphics[width=0.12\linewidth]{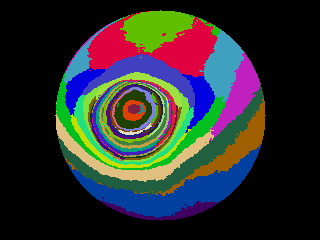}
}
\subfloat
{ 
\includegraphics[width=0.12\linewidth]{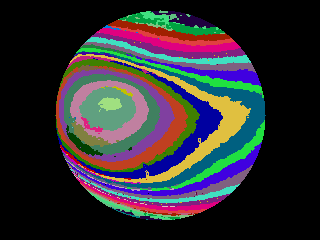}
}
\subfloat
{ 
\includegraphics[width=0.12\linewidth]{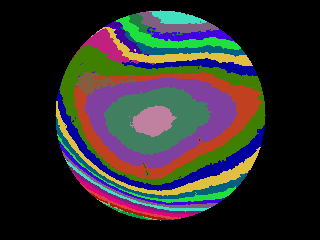}
}
\par
\setcounter{subfigure}{0}
\subfloat[cloth]
{ 
\includegraphics[width=0.12\linewidth]{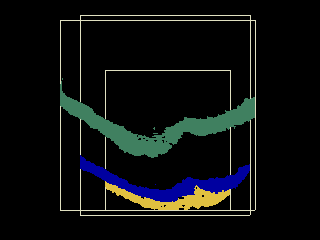}
}
\subfloat[feather]
{ 
\includegraphics[width=0.12\linewidth]{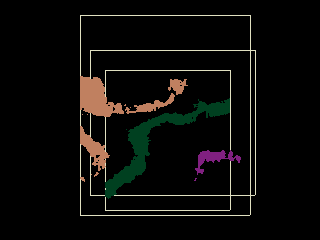}
}
\subfloat[glossy metal]
{ 
\includegraphics[width=0.12\linewidth]{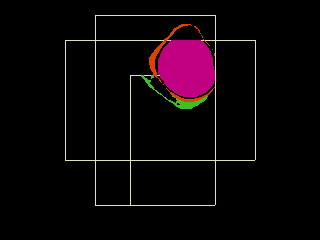}
}
\subfloat[paper]
{ 
\includegraphics[width=0.12\linewidth]{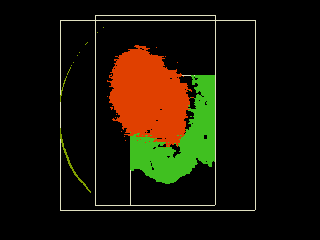}
}
\subfloat[sponge]
{ 
\includegraphics[width=0.12\linewidth]{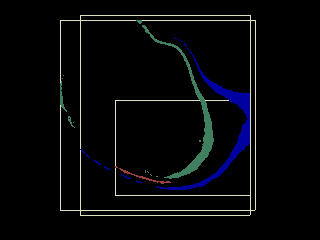}
}
\subfloat[wax paper]
{ 
\includegraphics[width=0.12\linewidth]{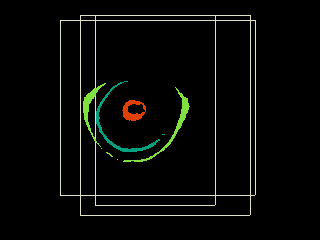}
}
\subfloat[wood]
{ 
\includegraphics[width=0.12\linewidth]{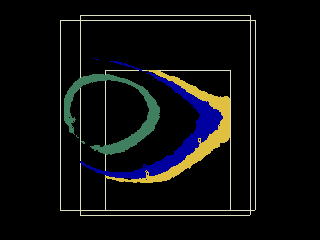}
}
\subfloat[wood]
{ 
\includegraphics[width=0.12\linewidth]{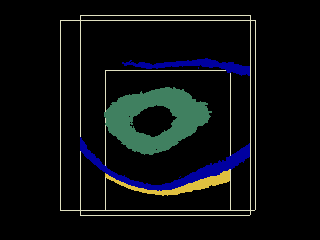}
}
\par
\end{center}
   \caption{Visualizing the reflectance-layout filters.The first row shows the reflectance disks for different materials. The second row shows the corresponding texton maps. The third row shows three of the dominant region-texton ($r,t$) combinations chosen for each class.}
\label{fig:textonvisual}
\end{figure*}

\section{Experiments}
\begin{figure}[t]
\begin{center}
\vspace{0.05in}
\includegraphics[width=3.1in]{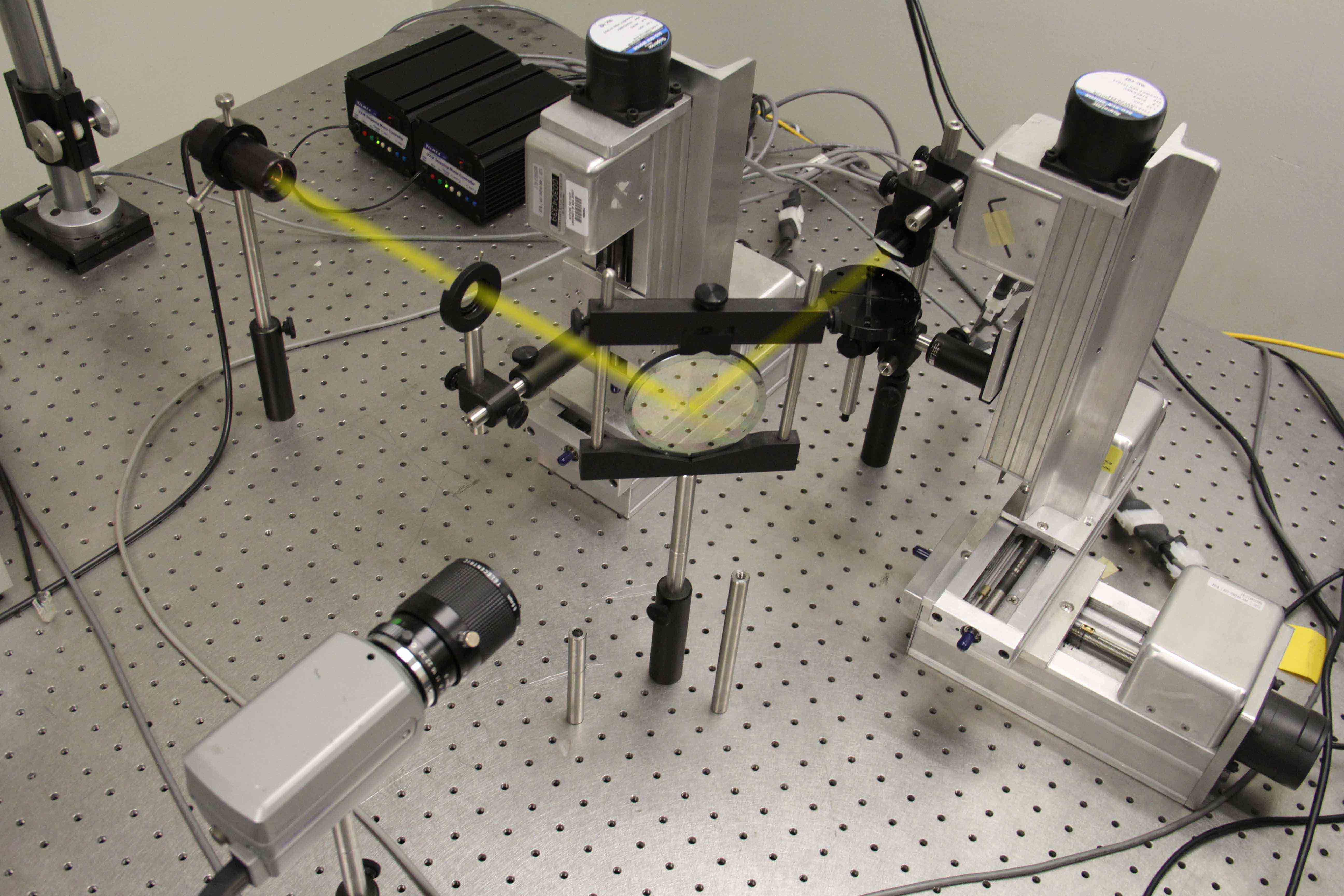}
\caption{Prototype of the camera for reflectance disk capture. The components follow the arrangement in Figure~\ref{fig:cameraschematic}. 
}
\label{fig:texcam}
\end{center}
\end{figure}

 \vspace{-0.04in}
{\bf Reflectance Disk Database } To evaluate the performance of our reflectance hashing framework,  a database of 3600 reflectance images is collected consisting of the following 20 surface classes: 
 cardboard, CD, cloth, feather, textured rubber, glossy ceramic, glossy metal, leather, linen, mattel metal, automotive painting, non-ferrous metal, paper, plastic, rock, rubber, sponge, velvet, glossy paper and wood.   The measurement camera  shown in Figure~\ref{fig:texcam} is arranged as in Figure~\ref{fig:cameraschematic} with a  video camera,  telecentric lens, light source, beam splitter and parabolic mirror.  The database includes 3 instances  per class, i.e.\ three different example surfaces per class,  10 spot samples per surface  with 3 illumination directions $(-10^\circ,0,10^\circ)$ where $0^\circ$ is frontal illumination aligned with the surface normal.  Additionally, two exposure settings are collected for each reflectance disk.  Therefore, the dataset consists of 180 reflectance disk images per class,  for a total dataset of 3600 reflectance disk images each of size $320 \times 240$.  This reflectance dataset is made publicly available (link provided in final publication).  
\begin{figure}[t]
\begin{center}
   \includegraphics[width=3.3in]{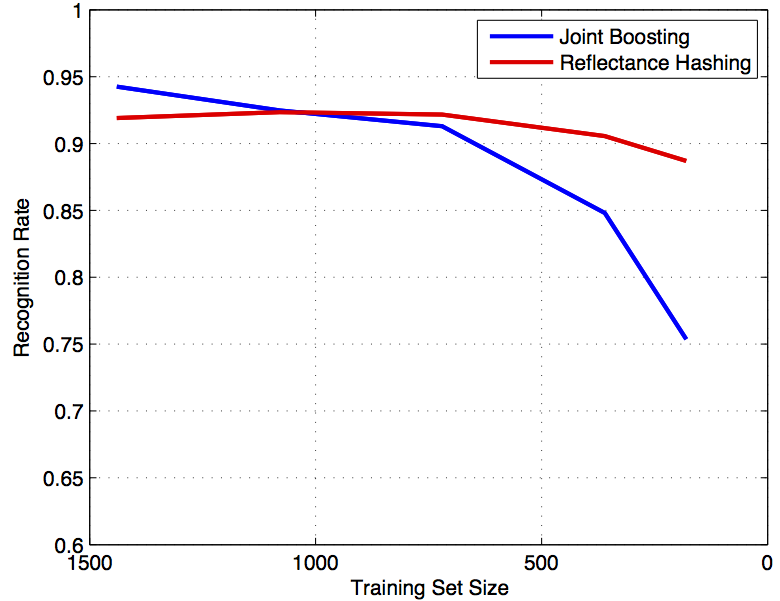}
\end{center}
   \caption{ Recognition rate as a function of training set size for both  boosting and reflectance hashing. Notice the performance of reflectance hashing is high even when the training set number is low.  }
\label{fig:errorwithsize}
\end{figure}
%
%
%

{\bf Material Recognition }
 We compare the performance of texton boosting with reflectance hashing for material classification.  We compared the classification result with a training set of $N$ random selected images  for both methods, and test the classification accuracy on a test set of $3600-N$ random selected images.  We vary $N$ from 360 $(10\%)$ to 1500. 
For the large set of training images, both methods perform well; however, the recognition rate of joint boosting decreases significantly, when decreasing the size of training set.  Reflectance hashing gives a more stable recognition rate as a function of the training set size as shown in  Figure \ref{fig:errorwithsize}.

\begin{figure*}[t]
\begin{center}
\includegraphics[width=6.6in,height=2.7in]{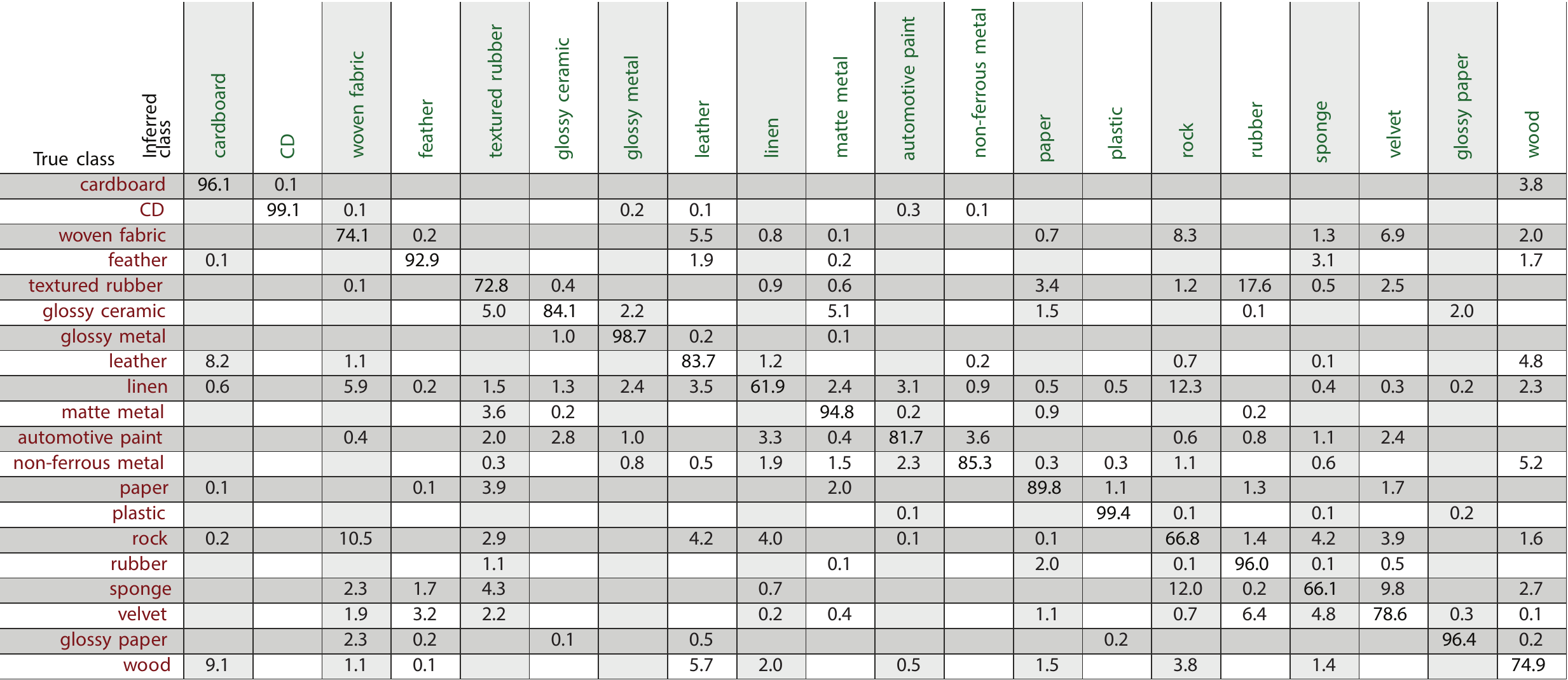}
\end{center}
\caption{
 Joint Boosting Classification Results. Confusion matrix with percentages row-normalized. Overall accuracy is 84.38\%.
}
\label{fig:confusion2}
\end{figure*}

\begin{figure*}[t]
\begin{center}
\includegraphics[width=6.6in,height=2.7in]{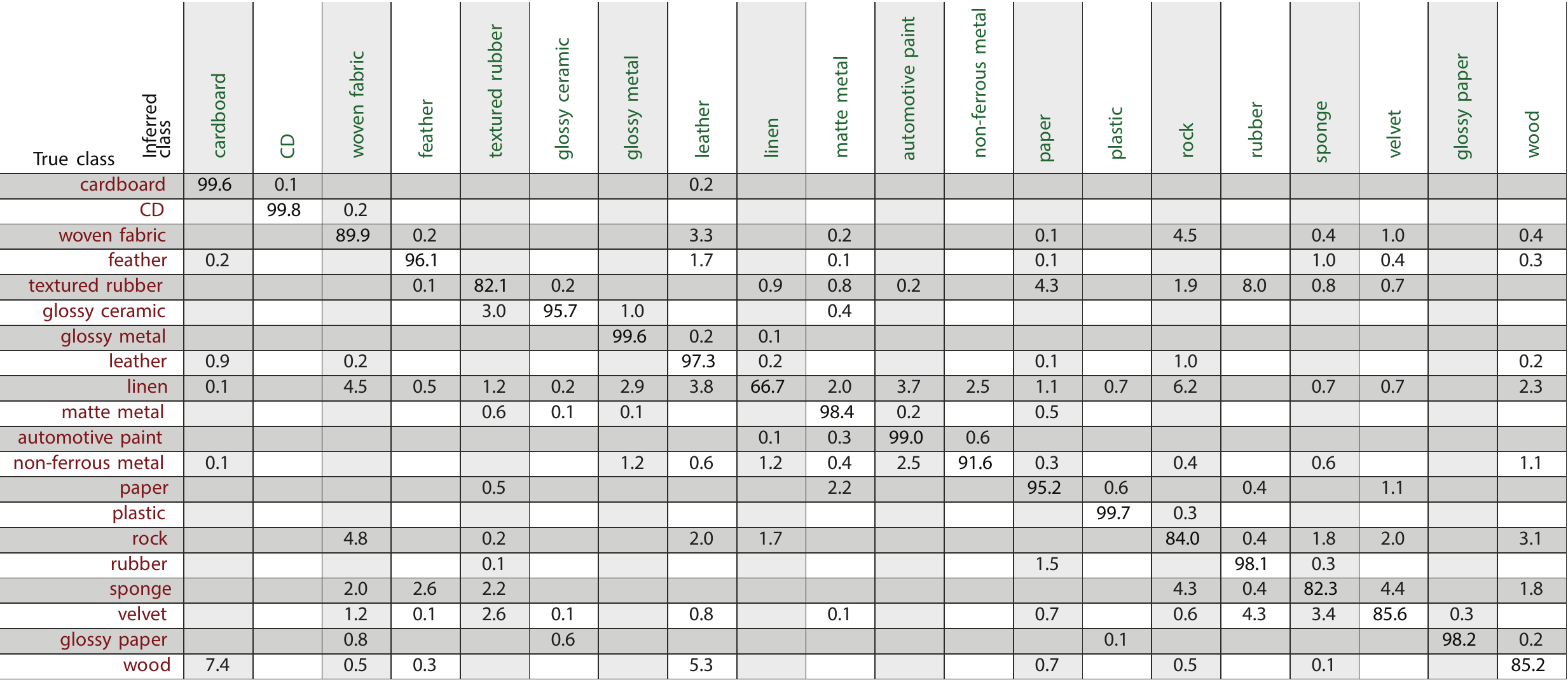}
\end{center}
\caption{
Reflectance Hashing Classification Results. Confusion matrix with percentages row-normalized. Overall accuracy is 92.3\%.
}
\label{fig:confusion1}
\end{figure*}

Figure \ref{fig:confusion2} shows the confusion matrix from $N=360$ classification with joint boosting with 700 iterations in the training stage, and  mean recognition rate of $84.38\%$. The parameter settings were $K=512$ textons and 200 subwindows or regions, with the random feature selection fraction $\tau =0.01$. 
Figure \ref{fig:confusion1} shows the confusion matrix obtained by our method of reflectance hashing, with the features selected in 700 iterations, and using iterative quantization \cite{Gong13b} as the binary embedding method.  The overall recognition rate is $92.3\%$, and several individual class recognition rates are significantly higher than the boosting method of Figure~\ref{fig:confusion2}.
For both methods, 5-fold cross validation was performed. 

We also make a baseline comparison with  texton histogram classification where no boosting is done and no region subwindows  are used. The histogram is computed over the entire reflectance disk.  
The recognition rate with 1500 training images is 79.53\% with a max rate of 100\% (plastic, glossy paper) and  a minimum of  45.6\% (sponge)
The recognition rate  
 with 360 training images is 41.94\% with a max rate of  89.2\% (cardboard) and a minimum of  18.3\% (linen). 

From the empirical results, we see that reflectance hashing provides  reliable recognition  even 
for a small training set.  This result has important implications for  real time material
recognition  since the approach may support  online training with compact training sets.

\begin{figure*}[]
\begin{center}
\subfloat[Nearest neighbor search precision]
{
\includegraphics[width=3in]{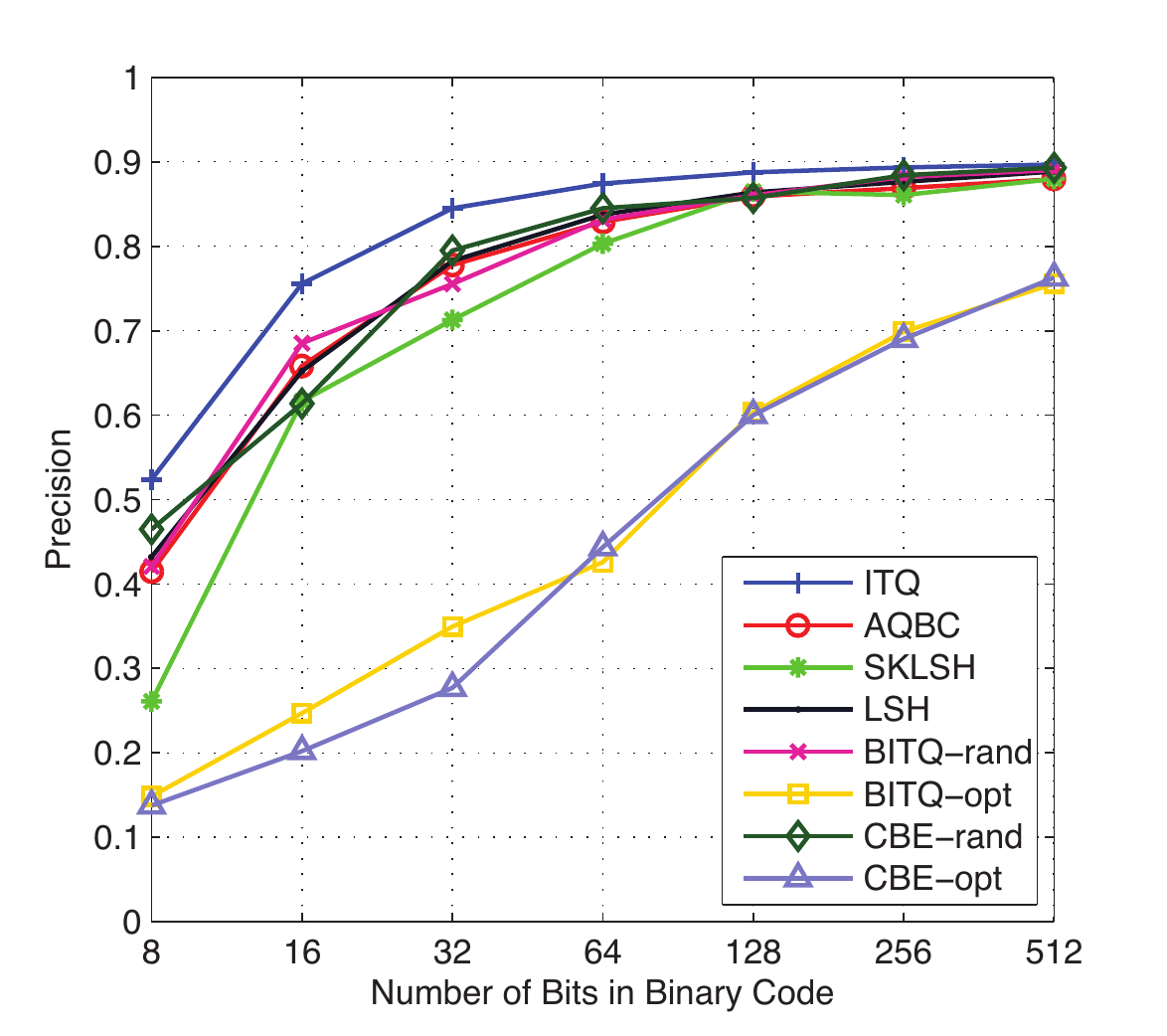}
}
\subfloat[Recognition Rate]
{
\includegraphics[width=3in]{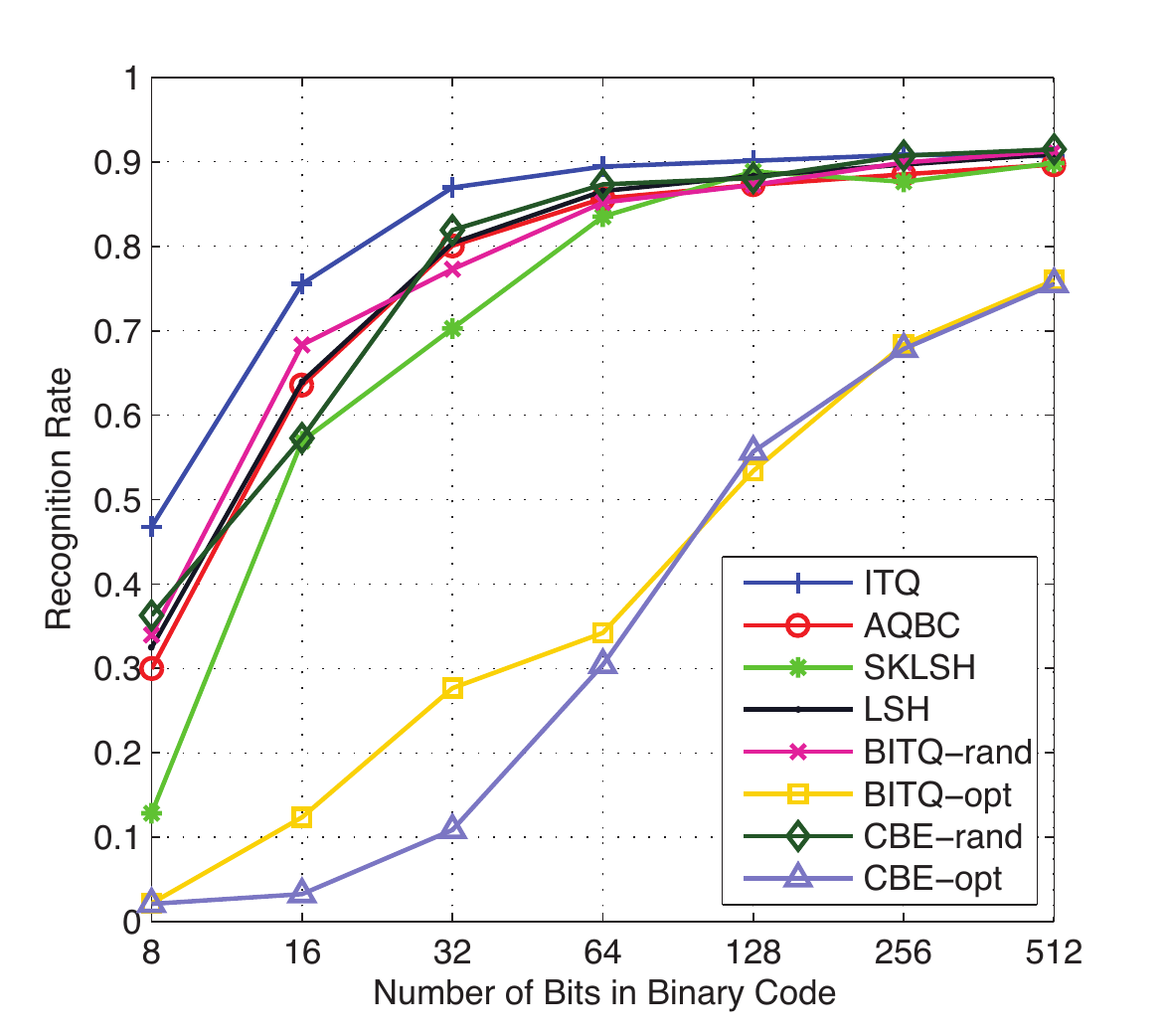}
}
\end{center}
\caption{Accuracy of nearest neighbor search and recognition rate as a function of the number of code bits for the binary embedding with 10 nearest neighbors.}
\label{fig:BinaryHashing}
\end{figure*}

{\bf Evaluation of Hash Codes }  We compare the performance of recent binary code methods for use in reflectance hashing.  These codes are use to represent the response from the  reflectance-layout filters as described in Section~\ref{sec:hashing}.  We test the following binary embedding methods: randomized (CBE-rand) and learned (CBE-opt) versions of circulant binary embedding (CBE) \cite{Yu14}, randomized (BITQ-rand) and learned (BITQ-opt) versions of bilinear embedding \cite{Gong13a}, iterative quantization (ITQ) \cite{Gong13b},  angular quantization-based binary codes(AQBC) \cite{Gong12},  locality-sensitive binary codes from shift-invariant kernels (SKLSH)   \cite{Raginsky09}, and a baseline method locality sensitive hash (LSH) \cite{Charikar02}.
We used the publicly available software implementation for these binary embedding  methods  \cite{Yu14}.
Recognition is  accomplished using a nearest neighbor search of binary codes with a Hamming distance metric and ten nearest neighbors. From Figure~\ref{fig:BinaryHashing}, we can see the binary embedding recognition rate reaches around $90\%$, when using 128 or 256 bits of binary codes.  The method of ITQ gives the best results for this material recognition task, and we use this binary code. The CBE-opt and BITQ-opt do not work well at low dimensional case, but the recognition rate increases quickly  with the number  of bits in the code. 

\section{Summary and Conclusion} 
We have introduced a novel framework for measurement and recognition of material classes. The approach encodes  high dimensional reflectance with a compact binary code.  We have compared several existing binary code methods to choose the most appropriate for this material recognition aproach.  The coding supports discrimination among the classes and can  be realized in embedded vision implementations.  Our method of reflectance hashing is compared to two popular baseline texton-based methods, boosting and texton histograms for material recognition. 
The results show excellent performance even for small training sets and provide a novel method based on reflectance for fast sensing and recognition of real-world materials. 
\vspace{-0.1in}
{\small
\bibliographystyle{ieee}
\bibliography{cvpr2015_zhangdanako.v5}
}

\end{document}